\documentclass[preprint,12pt]{elsarticle}

\usepackage{graphicx, subcaption}
\usepackage{amssymb}
\usepackage{algorithm}
\usepackage{algpseudocode}
\usepackage{tabularx}
\newtheorem{Def}{Definition}

\usepackage{lineno}
\usepackage[table]{xcolor}
\usepackage{pgf}
\usepackage{collcell}
\usepackage{booktabs}
\usepackage{fourier} 
\usepackage{array}
\usepackage{makecell}
\usepackage{amsmath}
\usepackage{csquotes}
\DeclareMathOperator*{\argmax}{arg\,max}

\newcommand*{\MinNegNumber}{-0.3}%
\newcommand*{\MaxNegNumber}{0}%
\newcommand*{\MinPosNumber}{0.000000001}%
\newcommand*{\MaxPosNumber}{0.52}%

\newcommand{\ApplyGradient}[1]{%
\ifdim #1 pt > -0.00001 pt
      \pgfmathsetmacro{\PercentColor}{100.0*(#1-\MinPosNumber)/(\MaxPosNumber-\MinPosNumber)}%
     \edef\x{\noexpand\cellcolor{green!\PercentColor}}\x\textcolor{black}{#1}%
\else
    \pgfmathsetmacro{\PercentColor}{100.0*(1-(#1-\MinNegNumber)/(\MaxNegNumber-\MinNegNumber))}%
     \edef\x{\noexpand\cellcolor{red!\PercentColor}}\x\textcolor{black}{#1}%
\fi
}
\newcolumntype{R}{>{\collectcell\ApplyGradient}{r}<{\endcollectcell}}

\journal{Neural Networks}

\begin{document}

\begin{frontmatter}


\title{Simplifying the explanation of deep neural networks with sufficient and necessary feature-sets: case of text classification}

\author[main_address,secondary_address]{Florentin Flambeau Jiechieu Kameni\corref{cor1}}
\cortext[cor1]{Corresponding author}
\address[main_address]{Department of Computer Science - University of Yaounde I, Yaounde, Cameroon}
\address[secondary_address]{IRD, UMMISCO, F-93143 Bondy, France}
\ead{florentin.jiechieu@facsciences-uy1.cm}
\ead{florentin.jiechieu@mintp.cm}

\author[main_address,secondary_address]{Norbert Tsopze}
\ead{norbert.tsopze@facsciences-uy1.cm}

\begin{abstract}
During the last decade, deep neural networks (DNN) have demonstrated impressive performances solving a wide range of problems in various domains such as medicine, finance, law, etc.  Despite their great performances, they have long been considered as black-box systems, providing good results without being able to explain them. However, the inability to explain a system decision presents a serious risk in critical domains such as medicine where people's lives are at stake. Several works have been done to uncover the inner reasoning of deep neural networks. Saliency methods explain model decisions by assigning weights to input features that reflect their contribution to the classifier decision. However, not all features are necessary to explain a model decision. In practice, classifiers might strongly rely on a subset of features that might be sufficient to explain a particular decision. The aim of this article is to propose a method to simplify the prediction explanation of One-Dimensional (1D) Convolutional Neural Networks (CNN) by identifying sufficient and necessary features-sets. We also propose an adaptation of Layer-wise Relevance Propagation for 1D-CNN. Experiments carried out on multiple datasets show that the distribution of relevance among features is similar to that obtained with a well known state of the art model. Moreover, the sufficient and necessary features extracted perceptually appear convincing to humans.
\end{abstract}


\end{frontmatter}


\section{Introduction}
\label{S:1}
In human understanding, a decision should be explained. Medical practitioners use diagnostic results to explain medical prescriptions to their patients, while lawyers explain court decisions using facts. Artificial intelligence models have evolved over the years, becoming more and more sophisticated and providing outstanding results in solving complex problems in a wide range of applications such as : text processing, image processing, medical analysis, etc. In the beginning, researchers were just concerned about improving the accuracy of artificial neural models. Therefore, deep neural networks (DNN) have emerged providing better results than the past models. Even though the results provided by DNN are satisfactory enough, they still suffer from the problem of opacity. They are often considered as black boxes, providing results without being able to explain them. In fact, entrusting important decisions to a system that cannot be explained presents obvious risks \cite{adadi2018peeking}. 

The problem of model interpretability has become a major concern in machine learning research communities since some machine learning programs were reported being racist, making social discrimination despite having good results in tests data \cite{Deep_explanation_survey, lowry1988blot}; other programs were found using inappropriate features (background of an image for example) to compute their decisions \cite{Deep_explanation_survey,MILLER_19}. 

Model explanation presents more than one benefit, among them : it eases the acceptability of artificial models in critical domains such as medicine, finance, etc.; it provides insights to improve learning algorithms and thus, models accuracy. Models explanation can also provide knowledge in the absence of domain theory.

Techniques to explain artificial models vary from one kind of model to another. While it is easy to interpret a decision tree or a model based on association rules for instance, the explanation of deep neural networks can be challenging. Several techniques reported in \cite{arrieta_et_al_2019,Deep_explanation_survey,MILLER_19, adadi2018peeking} have been proposed to explain deep neural networks :

\begin{enumerate}
\item Visualizing the network parameters: In images classification this method consists in visualizing the convolution filters and projecting them on the input images to highlight the regions of the images used for the classification

\item Inspecting the model: This method consists in providing a set of representations to understand how the model works and why it returns certain predictions more than others.

\item Using a saliency mask: this method consists in pointing out the cause of certain model outputs. It is very important in image or text classification because users can visualize the part of the input mainly responsible for the model outcome.

\item The sensibility analysis which consists in finding the influence of a given input feature to the model outcome.
\end{enumerate}

The present paper focuses on the local interpretation of CNN models built for text classification, which is the ability to explain or provide a meaningful and understandable justification to the model decision for a given input. The majority of works on CNN models explanation concerns image processing \cite{arrieta_et_al_2019,Deep_explanation_survey}. Only few research works have been conducted to interpret CNN predictions built for text classification \cite{jacovi_2018,Ribeiro_2016,zhang2015sensitivity,karpathy2015visualizing}. 

Previews studies have focused on the visualization of n-grams detected by filters. Thereafter, Jacovi et al. \cite{jacovi_2018} showed that some convolutional filters select accidental n-grams features because no other n-grams scored higher than them and those n-grams are not relevant in the explanation. However, even though certain n-grams would be considered accidental, they still influence the classifier decision. Moreover, their method lacks of genericity because it only applies to CNN with no hidden layer in the fully connected stage. 

In this article, we propose a method to explain CNN built for text classification using the principle of Layer-wise Relevance Propagation (LRP) \cite{Bach_LRP}, with a small adaptation of the contribution ratio formula which enables to clearly distinguish positive contributing n-gram features from negative ones. 

Indeed, positive contributing n-gram features are n-grams that contribute to increase the value of the predicted class and negative n-grams have the opposite effect. Therefore, positive n-grams are likely the features that might be useful to explain a classifier decision. However, as Miller suggested, not all features are necessary to explain a model decision \cite{MILLER_19}. 

Carter et al. \cite{carter2019made} have recently proposed to find the collection of Sufficient Input Subsets (\textbf{SIS}) which they defined as the minimal subset of features whose observed values alone suffice for the same decision to be reached, even if all other input feature values are missing. However, they do not distinguish positive features from negative ones. Indeed, the presence of only positive features without any negative ones might naturally suffice for the same decision to be reached. But in practice, inputs generally consists of both positive and negative features.  That is why we think that sufficient features must absolutely be determined in presence of all negative features. In that case, sufficient features would be subsets of positive features that in presence of negative ones suffice for the same decision to be reached.

In this study we refine the definition proposed by \cite{carter2019made} and propose the concept of Sufficient feature-sets which are subsets of positive contributing features such that the inhibition of the effect of all the other positive features does not change the classifier decision. We also propose to find necessary features which are positive features such that their inhibition changes the classifier decision. 

Indeed, necessary features if exists, represent key features (key n-grams responsible for the decision). On the other hand, sufficient feature-set can be used to simplify the explanation provided to human in the light that not all positive contributing features are not necessary to explain the model decision.

The main contributions of this work are :
\begin{enumerate}
\item We propose an LRP-based method to determine the relevance of n-grams features in a 1D-CNN architecture built for text classification ;
\item We also propose algorithms to evaluate sufficient and necessary feature sets in order to simplify the explanation, highlighting key features ;
\item We carried out multiple experiments in different applications of text classification including sentiment analysis and question answering to demonstrate the effectiveness our the method.
\end{enumerate}

The rest of this article is structured as follows: section 2 focuses on the presentation of some closely related works. Section 3 presents the principle of text classification using CNN; section 4 summarizes the LRP method. In section 5, our method to explain 1D-CNN is discussed, algorithms used to compute sufficient feature-sets and necessary features are also presented. The experiments and results are the aim of the sixth section.

\section{Related Works}
The explanation of deep learning models often presented as black box models has become a topical issue in recent years. Most of the works about model explanation reported in these surveys \cite{Tjoa2019_survey,Deep_explanation_survey} are concerned with image processing and they cannot easily be transposed to text processing due to the discrete nature of texts \cite{jacovi_2018}. Works aiming to interpret machine learning algorithm can be grouped in two major categories according to \cite{Tjoa2019_survey}: 
interpretability by mathematical structures which concerns the use of these structures to reveal the mechanism of machine learning and neural network algorithms \cite{zeiler2013visualizing,kim2018interpretability,tishby2015deep,shwartz2017opening,Selvaraju_17,LiuMA20}; and the perceptive algorithm which tends to present the user with explanatory elements that can be humanly perceived. The perceptive explainability includes : the saliency method, which explains the decision of an algorithm by assigning values that reflect the importance of input components based on their contribution to the classifier decision \cite{karpathy2015visualizing, ghorbani2019towards, shrikumar2016not, zintgraf2017visualizing, kindermans2017learning,plamen_20}; the signal method that observes the stimulation of neurons or a collection of neurons \cite{erhan2009visualizing, szegedy2015going, kindermans2019reliability}; and the verbal method in which verbal chunks or sentences are provided to the user as explanation \cite{caruana2015intelligible, caruana2015intelligible, letham2015interpretable}. In this section, we will focus on the saliency method as the explanation method proposed in this article falls into this category. 
    
LIME (Local Interpretable Model-Agnostic Explanations) \cite{Ribeiro_2016}  is a model independent interpretability method, in the sense that it is able to explain without needing to explore the inner functioning of the model. The principle of LIME method is to perturb the input around its neighborhood and observe the changes in the model predictions. For example, suppose we want to explain the prediction for the sentence \enquote{I hate this movie}. Lime will perturb that sentence and get the output predictions of resulting sentences such as : \enquote{I this movie}, \enquote{hate this movie}, \enquote{hate movie}, etc. From these perturbations, the relevancy of input words are determined. Even though LIME produces good results in practice, it does not \enquote{unblackbox} the model and the fidelity of the explanation to the exact inner model functioning can still be questioned. Another limit of LIME is that, to explain a single instance, it computes the outcome of multiple other instances obtained by varying the initial instance : this introduces a supplementary cost in terms of time complexity. 

Jacovi et al. \cite{jacovi_2018} have developed a method to understand how the CNN classifies a text. They examined the CNN parameters and showed that filters used in the convolution layer may capture several different semantic classes of n-gram by using different activation patterns : Informative n-grams selected by the pooling and used to classify the text, and, Uninformative n-gram eliminated by the pooling. They also distinguish between n-grams : deliberate and accidental n-grams. Deliberate n-grams are effectively informative with higher score regarding the final decision, accidental n-gram are n-gram selected by the pooling despite having a low score, because no other n-gram feature scored higher than them. However, their method applies only to a limited range of CNN architectures, those with only one layer in the fully connected stage. In addition, the fact that different convolution filters may select the same n-gram is not taken into account when scoring the n-gram features.


The Layer-wise relevance propagation was firstly introduced in  \cite{Bach_LRP}. In this method, the relevance of input features is evaluated by computing their contributions to the output of a particular neuron. The contributions are evaluated such that, the sum of contributions of units in a layer to a unit in the next layer is equal to the output of the latter. Other articles using this method includes \cite{lapuschkin2019unmasking, lapuschkin2019unmasking, samek2016evaluating, arras2017relevant}. In \cite{arras2017relevant} authors use LRP to explain CNN as for images. LRP is used to weight each component of the word embedding vectors as if they were pixels. The relevance of a word is obtained by summing up the relevance of each component of the corresponding word vector.



\section{Text classification using CNN}
\label{s:text-cnn}
\begin{figure}[ht]
\centering\includegraphics[width=1\linewidth]{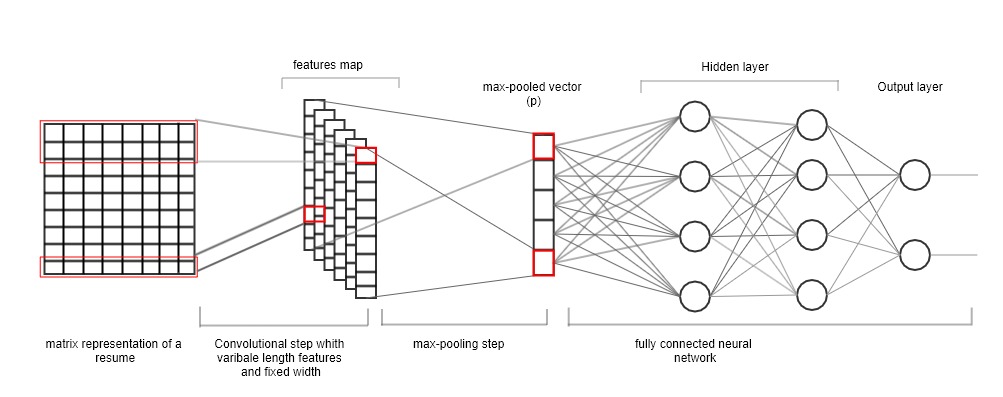}
\caption{Text classification using CNN}
\label{fig:archi_full}
\end{figure}

The architecture of the CNN described in this section is based on the architecture proposed by Kim \cite{kim-2014-convolutional} for sentence classification using CNN. Below we explain how the components of the CNN architecture work. Figure \ref{fig:archi_full} shows the different layers of the architecture. Each layer realizes specific operations.

\paragraph{The CNN Input layer}

The CNN takes as input a sequence of word embedding vectors representing the text to classify. A word $w$ can be modeled as a d-dimension vector of numerics : $w \in R^{d}$. Consequently, a text of length $n$ which is a sequence of $n$ words can be modeled as a matrix $M$ of dimension $n\times d$ ie $M={w_1,w2,...,w_n} \in R^{n\times d}$. An l-word n-gram $u_i = <w_i,..,w_{i+l-1}>\ (0 \leq i \leq n-l)$ is a sequence of $l$ consecutive words in the input text, $u_i \in R^{l\times d}$. 

\paragraph{1D-Convolution layer}
The convolution layer consists of 1D-convolution filters. A 1-kernel size convolution filter $f_j$ can be modeled as a $d-dimension$ vector (same shape as a word). The kernel size of a filter refers to the height of the filter sliding window since the width is fixed and corresponds to the word embedding dimension. A filter of kernel size 2 will slide vertically over two consecutive words at one time as shown in Figure \ref{fig:archi_full}. The convolution performs the scalar product $<u_i,f_j>$ between an n-gram $u_i$ and a filter $f_j$. The convolutions of a particular filter $f_j$ among the whole input matrix results in a column vector $F,_j$ called features map associated to $f_j$, and for the whole filters, the result is a matrix $F \in R^{(n-l+1)\times m}$ where $m$ is the total number of filters and the columns of the matrix represent single features maps associated to convolution filters.
The values inside the features map are rectified by setting all negative values to zero (Rectified Linear Unit) before going as input to the max-pooling layer. To summarize, the convolution layer can be modeled as a multivariate function $c:R^{n\times d} \rightarrow R^{(n-l+1)\times m}$ which takes as input a matrix of words embedding and returns a matrix of feature maps.

\paragraph{Max-pooling layer : Global Max-pooling}
The global max-pooling filter picks the maximum value in each rectified features map (each column of $F$), that is the value corresponding to the word which had the highest convolution score with the filter associated to that features map. The result of the global max-pooling is a $m-dimension$ vector. The max-pooling layer can then be modeled as function $mp:R^{(n-l+1)\times m} \rightarrow R^m$ which takes a matrix of features map and returns a vector containing the maximum value in each column. 

\paragraph{Fully connected neural network (FCNN)}
The FCNN takes as input the max-pooled vector $P$ (vector resulting from the max-pooling operation) and produces an output vector corresponding to the activation of each output unit. Therefore, the FCNN can be modeled as a function $h:R^m \rightarrow R^c$, $c$ being the number of output units. Let's denote by $L$ the total number of layers in the FCNN, $h^i$ the output vector of the $i$-$th$ layer, starting from  $h^0 = P$ (the max-pooled vector), and finishing with a special output layer $h^L$ which computes the output of the network. If $g$ is the activation function of hidden units, then, the activations of units in layer $k$ (in matrix-vector notation) is given by the equation \ref{eq:activation_k} where $b^k$ is the vector of biases and $W^{k}$ is the matrix of weights connecting layer $k-1$ to layer $k$.

\begin{equation}
    h^{k} = g(W^{k}h^{k-1} + b^k)
    \label{eq:activation_k}
\end{equation}

The activation of a single unit $i$ in layer $k$ is given by the Eq. \ref{eq:activation_ki}
\begin{equation}
    h^{k}_{i} = g(\sum_{j}{W^{k}_{ij}h^{k-1}_j} + b^{k}_i)
    \label{eq:activation_ki}
\end{equation}

Then the activations of the output units will be given by the Eq. \ref{eq:activation_output} (in matrix-vector notation):

\begin{equation}
    h^{L} = g(W^{L}h^{L-1} + b^L)
    \label{eq:activation_output}
\end{equation}

The 1D-CNN mathematically can be modeled as a function $f$ which realizes the composition of the functions c, m and h : $f = $ \( h \circ m \circ c\) .
\section{Layer-wise relevance propagation (LRP)}
\label{s:lrp}
LRP \cite{Bach_LRP} is a technique used to evaluate the contribution of input features to the neural network's output. The principle is to compute the contributions of units in the last hidden layer, then back-propagate them up to the input features. In this section, we will show how LRP is used to evaluate the contribution of the component of the max-pooled vector (see Figure \ref{fig:archi_full}). 

Let's denote by $f:R^l \rightarrow R^c$ the vector-value multivariate representing the mapping between the inputs features and the output of the network, where $l$ represents the size of the input vector (size of the max-pooled vector) and $c$ represents the number of output classes. We denote by $z^l_{ij} = w^{(l+1)}_{ij}h^{(l)}_i$ the product of the activation of the unit $i$ in layer $l$ and the synaptic weight connecting $i$ to the unit $j$ in layer $l+1$. 

We denote by $z^l_j = \sum_i{w^l_{ij}h^{(l-1)}_i + b^l_j} = \sum_i{z^l_{ij} + b^l_j}$ $(0<=l<L)$ the preactivation of the neuron $j$ in layer $l$. 

\begin{Def}
The contribution of a unit $i$ of the $l-th$ layer to a unit $j$ of the $(l+1)^{th}$ layer $(0<=l<L)$ denoted by $R^{(l,l+1)}_{ij}$ is the extend to which the unit $i$ has contributed to the preactivation value of the unit $j$. 
\end{Def}
Formally the contribution of the unit $i$ to the output unit $j$ is calculated using the equation \ref{eq:lrp-ratio} where $w^l_{ij}$ is the connection weight between the neuron $i$ of the layer $l$ and the unit $j$ of the layer $k+1$; and $h^l_i$ the activation of the unit $i$. 
\begin{equation}
\label{eq:lrp-ratio}
R^{(l,l+1)}_{ij} = \frac{z^l_{ij}}{z^{(l+1)}_j} = \frac{h^l_i*w^{(l+1)}_{ij}}{\sum_{k}{h^l_kw^{(l+1)}_{kj} + b^{(l+1)}_j}}
\end{equation}

To prevent $R^{(l,l+1)}_{ij}$ to take underbounded values due to potential small values of $z^{(l+1)}+1$ we will consider equation \ref{eq:emc3} instead, where $\epsilon$ is the stabilizer.
\begin{equation}
\label{eq:emc3}
R^{(l,l+1)}_{ij} = \left\{
    \begin{array}{ll}
        \frac{z^l_{ij}}{z^{(l+1)}_j+\epsilon} & \mbox{if } z^{(l+1)}_j \geq 0 \\
        \\
        \frac{z^l_{ij}}{z^{(l+1)}_j-\epsilon} & \mbox{if } z^{(l+1)}_j \leq 0.
    \end{array}
\right.
\end{equation}



Knowing the contribution of units of the $(l+1)-th$ layer  $(0\leq l<k-1)$ to the value predicted by the output neuron $j$, we can evaluate the contribution $R^{l}_{ij}$ of each neuron $i$ in layer $l$ to the value predicted by the output neuron $j$ by the Eq. \ref{eq:recurrent_relevance}.

\begin{equation}
\label{eq:recurrent_relevance}
R^{l}_{ij} = {\sum_{k}R^{(l,l+1)}_{ik}*R^{l}_{kj}}
\end{equation}
where $R^L_{jj}=f_j(x)$ and $R^0_{ij}$ represents the contribution of input feature $x_i$ to the output neuron $j$.

The mathematical foundations of LRP method are discussed in \cite{Bach_LRP}. The LRP algorithm operates as follows :

We start by computing the contribution of each unit $i$ in the (L-1)th layer to the value predicted for each unit $j$ in the output layer (L-th layer) using the equation \ref{eq:lrp_output_unit} 

\begin{equation}
\label{eq:lrp_output_unit}
R^{L-1}_{ij} = R^{(L-1,L)}_{ij}*f_j(x)
\end{equation}

$R^{(L-1,L)}$ is calculated using equation \ref{eq:emc3}. 

Thereafter, the contributions of units in layer $L-2$ to layer $L-1$ are calculated using the recurrent equation \ref{eq:recurrent_relevance}. 
The process is iterated until the contribution of input features to the outputs values predicted by the network are determined. 

LRP has widely been applied to image processing in order to weigh input pixels. But the discrete nature of text makes it difficult to apply in text CNN without any adaptation.

\section{Prediction Explanation}
\label{s:prediction-explanation}
The goal of the prediction explanation here is to highlight key features which have led to the classifier decision and that can be perceived by a human as a reason for that decision. 
As described in Sect. \ref{s:text-cnn}, each component of the max-pooled vector corresponds to the maximum convolution results of a particular filter with each words' vector or sequences of words' vectors of the input text depending on the filter kernel size used. Therefore, recovering the n-gram selected by a filter of kernel size $k$ simply consists in selecting the sequence of $k$-words which produced the maximum convolution result with that filter. Only those n-grams will further influence the classifier final decision \cite{jacovi_2018}. 

Based on the later assumption, the general principle of the explanation method described in this section consists in : (1) using the LRP approach to compute the relevance of each component of the max-pooled vector; (2) then recovering the n-grams associated to each of those components and determine their relevance. The method to explain the model prediction for an input text can be recapitulated as follows :
\begin{enumerate}
    \item First, a text is presented as input to the CNN and the outputs $f(x)$ is determined (Classification);
    \item Next, an adapted LRP that we will call LRP-A is used to compute the contribution of each component of the max-pooled vector to the output $f(x)$;
    \item Then, the n-gram features are determined and their relevance computed;
    \item After, sufficient and necessary n-gram features are computed;
    \item Finally, n-grams are classified into positive, negative sufficient or necessary. 
\end{enumerate}

The first step is described in detail in section \ref{s:text-cnn}. This section will focus on the prediction explanation method.

\subsection{LRP ratio Adaptation}
\label{s:lrp_ration_adaptation}
Sect. \ref{s:lrp} describes the the LRP method and its variants (LRP-$\epsilon$, LRP-0, LRP-$\alpha\beta$). However, we found a problem with either of these methods.  

\begin{figure}[ht]
\centering\includegraphics[width=0.5\linewidth]{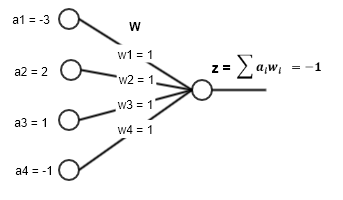}
\caption{Simplified neural network}\label{fig:negative-ratio}
\end{figure}

Let's consider the simplified neural network presented in Figure \ref{fig:negative-ratio}. Generally, activation functions in neural networks are increasing functions. This means that the greater the input of the function, the greater its output value. However, based on the architecture presented in Figure \ref{fig:negative-ratio}, the contribution ratio of the input unit $a_1$ to the linear sum of product $z$ will be $a_1w_1/\sum{a_iw_i= -3/-1 = 3}$, and the contribution of $a_2$, $a_3$ and $a_4$ will respectively be $-2$, $-1$, and $1$. This means that $a_1$ will be considered as the highest contributing input feature though it has a negative effect on the sum $z$ because it contributes to reduce the value of $z$, thus reducing the activation of the output unit. On the other hand, $a_2$ which actually contributes to increase the value of the sum of products $z$ will be considered having a negative contribution. Situations where the sum $z$ is negative, could distort the interpretation of some computed relevances when using Eq. \ref{eq:lrp-ratio}. To avoid that behavior, the contribution ratio of a unit $i$ is calculated by dividing the product $a_iw_i$ by the sum of absolute values of the products $z' = \sum{|a_jw_j|}$, which leads to Eq. \ref{eq:rectified_lrp}. Using Eq. \ref{eq:rectified_lrp}, the contributions of $a_1$, $a_2$, $a_3$ and $a_4$ will respectively be $-0.43$, $0.28$, $0.14$ and $-0.14$; which better reflects the effect of each input variable on the output unit pre-activation function. 
\begin{equation}
\label{eq:rectified_lrp}
R^{(l,l+1)}_{ij} = \frac{z^l_{ij}}{z^{(l+1)}_j} = \frac{a_i*w_i}{\sum_{k}{|a_kw_k|}}
\end{equation}

\subsection{Contribution of n-gram features}
\label{s:ngram-contrib}
Having the output of an input $x$, the first step in the explanation is to calculate the contribution of each component of the max-pooled vector using the LRP (Sect. \ref{s:lrp}) method. We end-up with a matrix representing the contribution of each filter to each class. After that, the next step is to compute the contribution of each n-gram to the output units.

We recall that each component of the max-pooled vector corresponds to the maximum convolution value of the filter with every word vector of the embedding input matrix. A filter $f_j$ selects an n-gram $u_i$ if the convolution of that filter with the given n-gram has produced the maximum value.  In addition, different filters may select the same n-gram for the same input text. Therefore, the contribution or the relevance of an n-gram $u_i$ noted $C_{u_i}$ is calculated as the sum of the contributions of filters ($R_f$) which select that n-gram (Eq. \ref{eq:ngram_contribution}) .   

\begin{equation}
    R_{u_i}=\sum_{f_j \in A_{i}}{R_{f_j}}
    \label{eq:ngram_contribution}
\end{equation}
$R_{f_j}$ is the relevance of the max-pooled value associated to the filter $f_j$ which also defines the relevance of that filter. $R_{f_j}$ is calculated using the LRP method described in Sect. \ref{s:lrp}
$A_i$ is the adjacency list associated to the n-gram $u_i$, which is the set of filters that selects the n-gram $u_i$. Formally, the n-gram selected by a filter $f_j$ is the n-gram $u_i^*$ which maximizes the scalar product $<u_i,f_j>$ ($1\leq i \leq n-l-1$, $1\leq j \leq m$). where $n$ is the number of words in the input text, $l$ the number of words of the n-gram and $m$ the total number of filters.  $u_i^*$ is determined by the formula \ref{eq:filter_select} \cite{jiechieu2020skills}.
\begin{equation}
    \label{eq:filter_select}
    u_{i^*}=\argmax_{u}(<u_i,f_j>), u_i \in W 
\end{equation}

With  (Eq. \ref{eq:ngram_contribution}) it is possible to evaluate the contribution of each n-gram to each output independently. We could then consider that : the greater the value of the contribution of an n-gram to the value predicted for an output neuron $j$, the more it is relevant to explain the output of that neuron. However, when using softmax activation in the output layer, the activation of a unit depends on the output values of the other units. In that case, the relevance of an n-gram for a target class must also take into account its contribution to other classes. Therefore it would be fair to refine the latter assumption to : a relevant n-gram to a given class $c_j$ is an n-gram which highly contributes to the target class $c_j$ and contributes little to other classes. Based on this assumption, the relevance of an n-gram $u_i$ with respect to an output class $c_j$ is the difference between the contribution of that n-gram to the class $c_j$ and the mean of its contributions to other classes. This means that : the more an n-gram contributes to the class $c_j$ and the less it contributes to the other classes, the more it is relevant to the target class $c_j$. 
\begin{equation}
    R^{'i}_j= R^i_j - \frac{\sum_{p\neq j}{R^i_p}}{k-1}
    \label{eq:rel_softmax}
\end{equation}
$k$ is the number of classes.
\begin{table}[ht]
\centering
\caption{Contribution of some n-grams to the sentiment predicted for the sentence : \textit{\enquote{great pocket pc phone combination}}}\label{table:contrib_table_1}
\begin{tabular}{ c R R R }
\toprule
\multicolumn{1}{c}{} & \multicolumn{1}{c}{NEG} & \multicolumn{1}{c}{POS} & \multicolumn{1}{c}{REL-POS} \\
great &  -0.0971  &  0.0822  & 0.1794\\
great pocket pc & -0.1132 & 0.0789 & 0.1922\\
phone combination &  0.0019  &  0.0014  &  -0.0005\\
\toprule
\end{tabular}
\end{table}
For example, table \ref{table:contrib_table_1} shows the contributions of some n-grams to the sentiment predicted by a sentiment analysis model for the sentence \textit{\enquote{great pocket pc phone combination}}. The model predicted the \enquote{positive} class (\textbf{POS}). From this table, we can observe that the contribution of the 3-gram \textit{\enquote{great pocket pc}} to the positive class (\textbf{POS}) was less than that of the 1-gram \textit{\enquote{great}}. However, we will consider the 3-gram \textit{\enquote{great pocket pc}} as more relevant to the \enquote{positive} class because it also strongly negatively contributes to the \enquote{negative} class (\textbf{NEG}), compared to the word \textit{great}.

For logistic output units, where the activation of a unit only depends on its output value we can consider that $R^{'i}_j=R^i_j$.   

\subsection{n-gram polarity}

\begin{Def}
Let $u_i$ be the n-gram identified by the filter $i$ and $R^i$ the contribution vector of $u_i$ to the output units, ie $R^i_j$ is the contribution of $u_i$ to the output class $c_j$. 
The class explained by $u_i$ is the class $c_j^*$ where $u_i$ contributes the most. Formally,

\begin{equation}
    c_{j^*}=\argmax_{j}(R^i_j), 2 \leq j \leq k \label{filter_select}
\end{equation}

$k$ being the total number of classes.

\end{Def}

\begin{Def}
A positive n-gram for a target class $c_j$ is an n-gram which mostly contributes to $c_j$ among all other classes. 

An n-gram with negative contribution will be called a negative n-gram; and an n-gram with null contribution does not influence in the outcome of the target output unit.
\end{Def}
Positive n-grams for a predicted class are n-grams which can likely be used to explain the prediction of the classifier because they actually contribute to increase the output value of the predicted class. But not all positive n-grams are necessary to explain the target class. We assume that only a subset of n-gram features is sufficient to explain a prediction. They are key n-gram features on which the model mainly base to compute its decision. 

\subsection{Sufficient features-set}
let $X = (x_1,x_2,...,x_n)$ an input data and  $f(X) = y$. Let $U = \{u_i\}_{i=1}^{i=n}$ the set of n-gram features selected by convolution filters, $U^+ \subset U$ and $U^- \subset U$ such that $U = U^+ \cup U^-$, the subsets of positive respectively negative n-gram features selected by convolution filters with regards to the classifier output decision.
\begin{Def}
A sufficient feature set is a subset $S \subset U^+$ of positive n-gram features such that the inhibition of all other positive features except $S$-features does not change the final classifier decision for an input text $x$. Inhibiting an n-gram feature consists in inhibiting all the filters that select this n-gram; and inhibiting a filter consists in setting the corresponding max-pooled value to zero or setting the synaptic weights connecting the max-pooled value associated to that filter to zero.
\end{Def}
A sufficient set of features $S$ is minimum if there is no other sufficient set of features $S'$ such that $S' \subset S$.
\begin{Def}
The relevance of a sufficient feature-set $S$ with respect to a target class $c$, is the sum of the contributions of the n-gram features of $S$ to class $c$.
\end{Def}
We are interested in sufficient features-set with maximum relevance. Therefore, a greedy-like method can be used to find a minimum sufficient features-set with highest relevance. The aim of finding sufficient feature-sets is to simplify the explanation by providing to the user key features on which the model mainly relies on to produce its outcome. The algorithm \ref{alg:SF} is a greedy-like algorithm to determine sufficient features set.

\begin{algorithm}
\caption{SF-set : Get Sufficient features-set}\label{alg:SF}
\begin{algorithmic}[1]
	\Require A 1D-CNN model $f$ ; An n-words input sequence X;
	\Ensure S = A minimum sufficient features-set with maximum relevance
    \State $p = f(X)$ \Comment{The class predicted for the input X}
    \State $C = LRP(f,X)$ \Comment{The contributions of input features of the fully connected neural network to the output units}
    \State $F = f.Conv1D(X)$ \Comment{The intermediate result (Features map) of the 1D-convolution layer}
    \State $U = \argmax_j{F_{ij}}$ \Comment{Indices of n-gram features selected by convolution filters}
    \State $A = []$ \Comment{Dictionary containing the list of filters that select each n-gram feature}
    \State NC = vector of n-gram contributions, initialized to null. 
    \For{\texttt{i from 1 to m }}
        \State $A[U[i]] = A[U[i]] + \{i\}$\Comment{add filter i into the adjacency list related to n-gram U[i]}
        \State $NC[U[i]] = NC[U[i]] + C[i]$ \Comment{C[i] is the contribution of filter i}
    \EndFor
    \If{\texttt{ the output activation of $f$ is logistic}}
            \State $U^+ = \{ u \in U\ |\ C^i_p >0 \}$\Comment{p is the indice of the predicted class}
    \ElsIf{\texttt{output activation of $f$ is softmax}}
        \State $U^+ = \{ u \in U\ |\ C^i_p - \frac{\sum_{j\neq p}{C^i_j}}{|C^i|-1} > 0 \}$
    \EndIf
    \State $sort\ U^+\ elements\ in\ ascending\ order\ of\ their\ contribution$
    \State $y = p,\ i = 0,\ S = U^+$
    \While{\texttt{p == y}}
        \State $L = A[U[i]]$ \Comment{List of filters selecting U[i]}
        \State $f'$ = the model obtained by setting in f the weights connecting the max-pooled value corresponding to filters in L to zero
        \State $y = f'(X)$
        \If{\texttt{y == P}}
            \State Remove $U[i]$ from $S$ \Comment{U[i] is not necessary}
        \EndIf
        \State $i = i + 1$
    \EndWhile
\State \textbf{return} $S$ 
\end{algorithmic}
\end{algorithm}


\subsection{Necessary features}
let $X = (x_1,x_2,...,x_n)$ an input data and  $f(X) = y$. Let $U = \{u_i\}_{i=1}^{i=n}$ the set of n-gram features selected by convolution filters, $U^+ \subset U$ and $U^- \subset U$ such that $U = U^+ \cup U^-$, the subset of positive respectively negative n-gram features selected by convolution filters with regards to the classifier output decision.
\begin{Def}
A necessary n-gram feature is a positive n-gram $u \in U^+$ such that the inhibition of filters that select $u$ will change the classifier decision for the input text $X$. 
\end{Def}

The necessary feature-set is the set of necessary features. Necessary features present a major importance in explanation because according to the model, they were required to produce the decision given by the model. Algorithm \ref{alg:NF} is a greedy-like algorithm to determine the set of necessary features.  

\begin{algorithm}
\caption{NF-set: Get Necessary features-set}\label{alg:NF}
\begin{algorithmic}[1]
	\Require A 1D-CNN model $f$; an n-words input sequence X;
	\Ensure N = a set of necessary n-gram features
    \State $p = f(X)$ \Comment{The class predicted for input X}
    \State $C = LRP(f,X)$ \Comment{The contributions of input features of the fully connected neural network to the output units}
    \State $F = f.Conv1D(X)$ \Comment{The intermediate result (Features map) of the 1D-convolution layer}
    \State $U = \argmax_j{F_{ij}}$ \Comment{Indices of n-gram features selected by convolution filters}
    \State $A = []$ \Comment{ Dictionary containing the list of filters that select each n-gram feature}
    \State NC = vector of n-gram contributions, initialized to null. 
    \For{\texttt{i from 1 to m }}
        \State $A[U[i]] = A[U[i]] + \{i\}$
        \State $NC[U[i]] = NC[U[i]] + C[i]$ 
    \EndFor
    \If{\texttt{output activation of $f$ is logistic}}
            \State $U^+ = \{ u \in U\ |\ C^i_p >0 \}$\Comment{p is the indice of the predicted class}
    \ElsIf{\texttt{output activation of $f$ is softmax}}
        \State $U^+ = \{ u \in U\ |\ C^i_p - \frac{\sum_{j\neq p}{C^i_j}}{|C^i|-1} > 0 \}$
    \EndIf
    \State $sort\ U^+\ elements\ in\ descending\ order\ of\ their\ contribution$
    \State $y = p,\ i = 0,\ N = \{\}$

    \While{\texttt{$p \neq y$}}
        \State $L = A[U[i]]$
        \State $f'$ = the model obtained by setting in f the weights connecting the max-pooled value corresponding to filters in L of the FCNN input vector to zero
        \State $y = f'(X)$
        \If{\texttt{$y \neq p$}}
            \State $N = N + {U[i}$ \Comment{U[i] is necessary}
        \EndIf
        \State $i = i + 1$
    \EndWhile
\State \textbf{return} $N$ 
\end{algorithmic}
\end{algorithm}

\section{Experiments and Results}
The experiments\footnote{https://github.com/florex/xai-cnn-lrp} have been conducted on multiple channel CNN models based on Kim architecture \cite{kim-2014-convolutional} plus a trainable embedding layer. The explanation method has been tested on architectures consisting of 1 to 3 channels with filters of kernel sizes 1, 2 or 3. The word embedding layer consists of 50-dimension word vectors and the input texts were padded to a maximum length equals to 50.

\subsection{Datasets}
\label{s:dataset-description}
The method has been experimented on 4 datasets :
\begin{itemize}
    \item \textbf{IMDB : }A movie review dataset for binary sentiment classification \cite{maas-EtAl:2011:ACL-HLT2011};
    \item \textbf{sentiment140 : } A dataset containing 1,600,000 tweets extracted using the twitter api for the sentiment analysis task \cite{go2009twitter};
    \item \textbf{TREC-QA\_5500 : } A dataset for Question Answering Track with 5500 train samples \cite{qa_trac}
    \item \textbf{TREC-QA\_1000 : } A dataset for Question Answering Track with 1000 train samples \cite{qa_trac};
    \item Few already pretraited training data from \textbf{Imdb}, \textbf{Amazon} and \textbf{yelp} were also provided by \cite{kotzias2015group}.
\end{itemize}

Table \ref{table:dataset-classes} describes the classes of the above mentioned datasets.
\begin{table}[ht]
\caption{Dataset classes description}\label{table:dataset-classes}
\centering
\begin{tabular}{|c|c|c|}
\hline
\thead{dataset name} & \thead{number of\\ classes} & \thead{Classes description} \\
\hline
IMDB & 2 & \makecell{0 = Negative\\1 = positive} \\
\hline
sent140 & 3 & \makecell{0 = Negative\\2 = Neutral\\4 = Positive}\\
\hline
TREC-QA & 6 & \makecell{DESC=DESCRIPTION\\ENTY=ENTITY\\ HUM=HUMAN\\NUM=NUMBER\\LOC=LOCATION\\ABBR=ABBREVIATION} \\
\hline
\end{tabular}
\end{table}

A model was built for each task described above. Table \ref{table:testing-models} shows a summary description of models used to test the explanation method described in this article. The model name is related to the name of the dataset used to train the model.

\begin{table}[ht]
\caption{Description of models built to test the explanation method}\label{table:testing-models}
\centering
\begin{tabular}{|c|c|c|c|c|}
\hline
\thead{model name} & \thead{number of\\ channels} & \thead{kernel sizes\\per channel} & \thead{filters per\\channel} & \thead{model\\accuracy}\\
\hline
IMDB & 3 & [1,2,3] & [40,40,40] & 83.6\%\\
\hline
sent140 & 3 & [1,2,3] & [40,40,40] & 88,9\%\\
\hline
TREC-QA\_5500\_1ch & 1 & [1]& [40] & 78.14\%\\
\hline
TREC-QA\_5500\_3ch & 3 & [1,2,3] & [40,40,40] & 84.52\%\\
\hline
TREC-QA\_1000\_3ch & 3 & [1,2,3] & [40,40,40] & 74.80\%\\
\hline
\end{tabular}
\end{table}

Table \ref{table:testing-models} shows that the model built for sentiment classification using \textbf{Imdb} dataset has 3 channels with filters of kernel size respectively of 1, 2, 3 and each of these  channels has 40 filters; this makes a total of 120 filters in the convolution layer.

\subsection{Results and Discussions}

\begin{figure}[ht]
\begin{subfigure}{0.5\textwidth}
\includegraphics[width=1\linewidth, height=5cm]{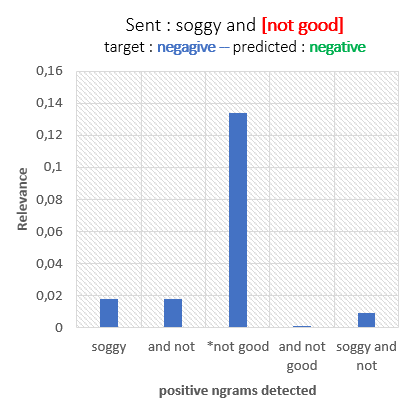} 
\caption{A sample sentence from sent140}
\label{fig:sent140-1}
\end{subfigure}
\begin{subfigure}{0.5\textwidth}
\includegraphics[width=1\linewidth, height=5cm]{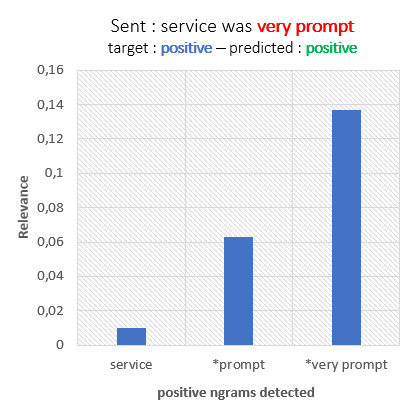}
\caption{A sample sentence from sent140}
\label{fig:sent140-2}
\end{subfigure}
\begin{subfigure}{0.5\textwidth}
\includegraphics[width=1\linewidth, height=5cm]{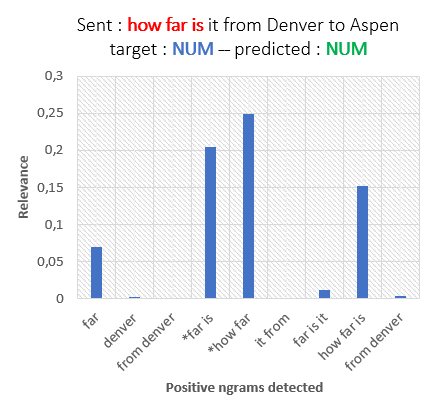}
\caption{A sample sentence from TREC-QA\_5500}
\label{fig:trec}
\end{subfigure}
\begin{subfigure}{0.5\textwidth}
\includegraphics[width=1\linewidth, height=5cm]{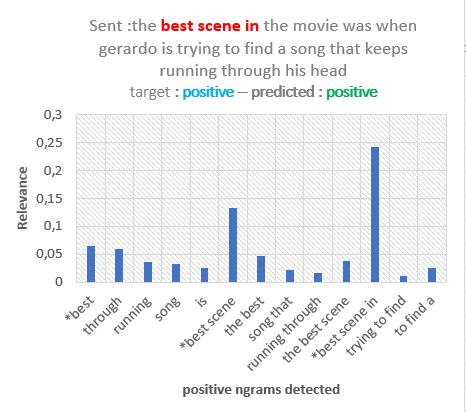}
\caption{A sample sentence from IMDB}
\label{fig:imdb}
\end{subfigure}
 \caption{Relevance of positive n-gram detected by filters.}\label{fig:positve_n-gram}
\end{figure}
The explanation method was tested on models described in table \ref{table:testing-models}. Figure \ref{fig:positve_n-gram} shows the distribution of relevance among positive n-grams for random sentences picked from the datasets described in Sect. \ref{s:dataset-description}. Labels with asterix (*) represent sufficient features sets detected by algorithm \ref{alg:SF}. They are also highlighted in red in the original sentence. Necessary features are terms inside the brackets. In Figure \ref{fig:trec} the model classified the question according to the type of the expected answer. Indeed, the sentence \textit{\enquote{how far is it from Denver to Aspen?}} is classified as \textbf{\enquote{NUM}} because the expected answer is a number representing the distance between Denver and Aspen. Several positive contributing n-grams \textit{(far, denver, from denver, ...)} were identified, but only the n-grams \textit{\enquote{far is}} and \textit{\enquote{how far is}} were sufficient to produce the same decision. This means that if all the filters that select other positive n-grams were inhibited except those selecting \textit{\enquote{far is}} and \textit{\enquote{how far is}}, the decision would have been the same. Likewise, Figure \ref{fig:imdb} has multiple positive n-gram features, however only three n-gram features contributed enough such that they were sufficient to produce the final decision according to the model. They are : \textit{\enquote{best}, \enquote{best scene}, \enquote{best scene in}}; and when projected on the sentence it gives \textit{\enquote{the \underline{best scene in} the movie was when gerardo is...}}, which greatly simplifies human perception of the explanation provided. 

In Figure \ref{fig:sent140-1}, the n-gram \textit{\enquote{not good}} is at the same time sufficient and necessary, which means that it has greatly contributed to classify the sentence \textit{soggy and [not good]} as \textit{NEGATIVE}.


\begin{table}[ht]
\centering
\caption{Relevance of n-grams to the class predicted for the question : \textit{\enquote{who was the star witness at the senate watergate hearings?}}}\label{table:contribution_table}
\begin{tabular}{ c R R R R R R R}
\toprule
\multicolumn{1}{c} {} & \multicolumn{1}{c} {DESC} & \multicolumn{1}{c} {ENTY} & \multicolumn{1}{c} {ABBR} & \multicolumn{1}{c} {HUM} & \multicolumn{1}{c} {NUM} & \multicolumn{1}{c} {LOC} & \multicolumn{1}{c} {REL-HUM}\\

senate &  0.02  &  -0.02  & -0.00 & -0.00 & 0.01 & -0.00 & -0.00\\
witeness & 0.01 & -0.04 & -0.01 & -0.00 & 0.04 & -0.01  &  -0.00\\
hearings &  0.03  &  0.03  &  0.02  &  0.00  &  -0.03  &  -0.03  &  0.01\\
water-gate &  0.01  &  0.01  &  0.01  &  0.00  &  -0.02  &  -0.03  &  0.01\\
who &  -0.15  &  0.07  &  -0.17  &  0.37  &  -0.28  &  -0.16  &  0.51\\
at &  -0.01  &  0.01  &  0.00  &  -0.00  &  -0.01  &  -0.00  &  -0.00\\
the &  0.00  &  -0.03  &  -0.07  &  0.08  &  -0.02  &  -0.02  &  0.11\\
was &  -0.01  &  -0.02  &  -0.02  &  0.04  &  0.00  &  -0.01  &  0.05\\
\toprule
\end{tabular}
\end{table}

Table \ref{table:contribution_table} shows the heat map of a one-channel model using the TREC-QA dataset (TREC-QA\_5500\_1ch model described in table \ref{table:testing-models}). Cells represent contributions of words labeling table rows to classes labeling table columns. The intensity of a cell indicates the degree of the contribution of the corresponding word to the corresponding class. The dark red color indicates a strong negative contribution while the dark green color indicates a strong positive contribution to the class and the white color indicates almost null contribution. The last column (REL-HUM) represents the relevance to the predicted class (\textbf{HUM}) as the difference between the contribution of a word to the class \textbf{HUM} and the mean of its contribution to the other classes as described in Sect. \ref{s:prediction-explanation} Eq. \ref{eq:rel_softmax}. From this table, we can figure out that the term \textit{\enquote{who}} does not only contribute highly to the class \textbf{HUM} but also contributes negatively to other classes except to \textbf{ENTY} where it contributes positively. By observing the relevance of each word to the predicted class, we can deduce that \textit{\enquote{who was the}} are the main terms that were responsible for the prediction of the class \textbf{HUM}, which means that the expected type of the answer to that question is \textbf{HUMAN}. 

\subsection{Discussion on sufficient and necessary set}
\begin{figure}[ht]
\centering\includegraphics[width=1\linewidth]{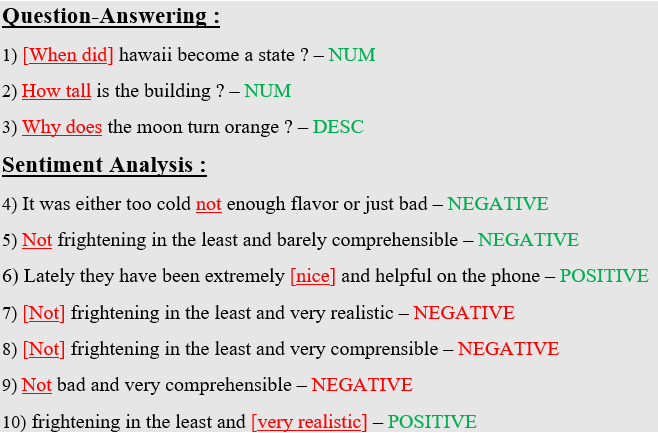}
\caption{Sufficient and necessary features}\label{fig:sufficient-featuers}
\end{figure}
Figure \ref{fig:sufficient-featuers} shows a list of sentences taken from a question answering dataset (\textbf{TREC-QA}) and from sentiment analysis datasets (\textbf{imdb}, \textbf{sent140}).  In each sentence the value after the dashes \enquote{--} represents the class predicted by the model. When this value is colored in green, it means that the model has correctly predicted the class, else the prediction is incorrect. The words colored in red in each sentence represent the sufficient n-gram features and those inside brackets represent necessary n-gram features. In sentence 1, the 2-gram \textit{\enquote{When did}} was at the same time sufficient and necessary to predict the class \textbf{NUMBER} as the expected type of the answer to that question. In sentence 2, \textit{\enquote{how tall}} was considered sufficient to predict the class \textbf{NUMBER} and there is no necessary feature.  

In sentence 5, the term \textit{\enquote{not}} was considered by the model as sufficient to predict the sentiment \textbf{NEGATIVE}. We performed little variations on that latter sentence, to change the underlying sentiment to POSITIVE, which have led to sentences 7, 8, 9, 10. We observe that the model has still predicted the sentiment NEGATIVE (for sentences 7, 8, 9) while still relying on the term \textit{\enquote{not}}. However, when the term \textit{\enquote{not}} is removed (sentence 10) the sentiment changes to POSITIVE, and the n-gram \enquote{very realistic} is determined as both sufficient and necessary to explain this new output. These examples show that in the case of sentences 7, 8, 9 the model has wrongly relied on the term \textit{\enquote{not}} to produce its decision. This is most likely due to insufficient training samples. Indeed, the model was built with a training set of 2250 sentences. A solution track would be to add in the training set, sentences which contain the term \textit{\enquote{not}} with similar contexts to those of sentences where the prediction was marked as incorrect.

\subsection{Comparison with LIME}
Since LIME is a statistical model agnostic method for explanation which also evaluates the relevance of each input feature to the output classes, we thought of comparing the distribution of relevance using our method with that obtained using LIME.
\begin{figure}[ht]
\begin{subfigure}{0.5\textwidth}
\includegraphics[width=1\linewidth, height=8cm]{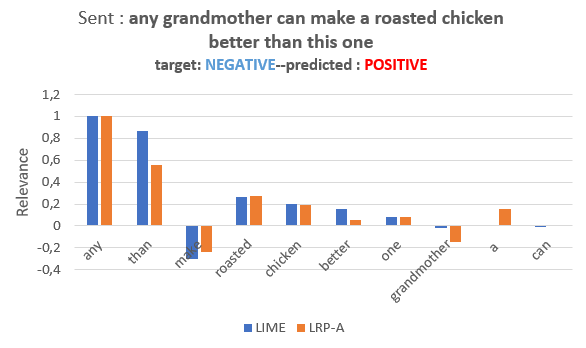} 
\caption{LIME and LRP-A on Sentiment Analysis}
\label{fig:imdb_dist}
\end{subfigure}
\begin{subfigure}{0.5\textwidth}
\includegraphics[width=1\linewidth, height=8cm]{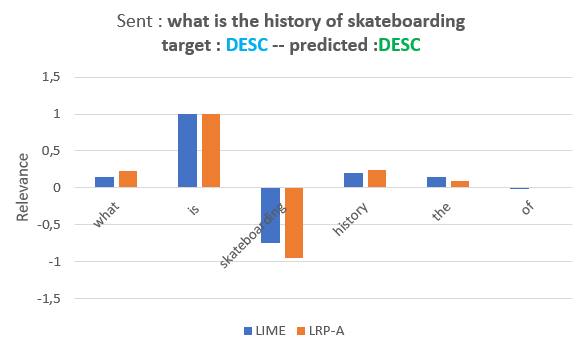}
\caption{LIME and LRP-A on TREC-QA dataset}
\label{fig:trec_dist}
\end{subfigure}
 \caption{LIME versus CLRP}\label{lime-vs-lrp}
\end{figure}
Figure \ref{lime-vs-lrp} shows the distributions of relevance using LIME (in blue) and using our method (LRP-A) for two sentences. The first sentence (Figure \ref{fig:imdb_dist}) is from \textbf{imdb} corpus and the second one (Figure \ref{fig:trec_dist}) is from the \textbf{TREC-QA} corpus. Datas have been normalized using Max-normalization in order to have both contributions in a comparable scale and to preserve the sign of each contribution. We observe that LIME and LRP-A (our method) have almost the same distribution of relevance among words for both sentences. The most important terms highlighted by both methods are the same. 

However, we can observe that no filter has selected the term \textit{\enquote{can}} with our method but LIME has still scored it though with low relevance. Since our method is a white box method, it is easy to verify that the term \textit{\enquote{can}} had no impact on the output since it was not selected by any max-pooling filter. Another interesting fact to notice is that in Figure \ref{fig:imdb_dist}, the model failed to predict the right class. Indeed, the model predicted the class \textbf{POSITIVE} while the expected class was the \textbf{NEGATIVE} class. When observing the distribution of relevance of words to the predicted class (\textbf{POSITIVE}) we can figure out that the model mainly relied on the features \textit{(any, than)} which cannot be perceived as a correct justification for that prediction. Therefore, we can deduce that the model relied on wrong features to predict the class POSITIVE. This is likely due to an insufficient number of samples when training the model. This knowledge could provide insights to improve the model accuracy. For example, by providing negative samples where the terms \textit{(any, than)} appear to tell the model not to associate them with the class \textbf{POSITIVE}.

\subsection{Impact of the LRP adaptation}
\begin{figure}[ht]
\centering\includegraphics[width=1\linewidth]{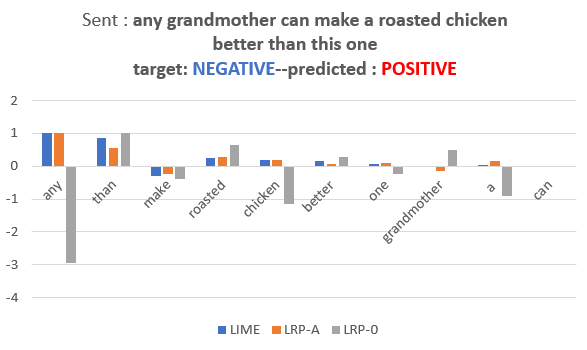}
\caption{Impact of the new contribution ratio formula}\label{fig:lime_clrp_lrp}
\end{figure}
As described in Section \ref{s:lrp_ration_adaptation}, to prevent undesired effects when the denominator in the LRP ratio (Eq. \ref{eq:lrp-ratio}) is negative, we proposed to use Eq. \ref{eq:rectified_lrp} to compute the ratio. Figure \ref{fig:lime_clrp_lrp} presents a comparison between the distribution obtained using LIME, the standard LRP-0 and LRP-A. We observe that while \text{LIME} and \text{LRP-A} have almost the same distribution, the sign of the contribution of certain words when using \text{LRP-0} diverge. This confirms the analysis done so far in section \ref{s:lrp_ration_adaptation}





\section{Conclusion}
\label{S:Conclusion}
We have described in this work, a method to explain 1D-Convolutional neural networks built for text classification. Experiments carried out on various datasets have shown that the distribution of relevance among the features matches with that obtained with a well-known statistical model agnostic method. In addition, the identification of sufficient and necessary features were shown to greatly simplify the explanation by highlighting key features. Techniques developed in this work can also be easily adapted to work with image processing using CNN.






\bibliographystyle{elsarticle-num-names}
\bibliography{sample.bib}

\begin{thebibliography}{37}
\expandafter\ifx\csname natexlab\endcsname\relax\def\natexlab#1{#1}\fi
\providecommand{\url}[1]{\texttt{#1}}
\providecommand{\href}[2]{#2}
\providecommand{\path}[1]{#1}
\providecommand{\DOIprefix}{doi:}
\providecommand{\ArXivprefix}{arXiv:}
\providecommand{\URLprefix}{URL: }
\providecommand{\Pubmedprefix}{pmid:}
\providecommand{\doi}[1]{\href{http://dx.doi.org/#1}{\path{#1}}}
\providecommand{\Pubmed}[1]{\href{pmid:#1}{\path{#1}}}
\providecommand{\bibinfo}[2]{#2}
\ifx\xfnm\relax \def\xfnm[#1]{\unskip,\space#1}\fi
\bibitem[{Adadi and Berrada(2018)}]{adadi2018peeking}
\bibinfo{author}{A.~Adadi}, \bibinfo{author}{M.~Berrada},
\newblock \bibinfo{title}{Peeking inside the black-box: A survey on explainable
  artificial intelligence (xai)},
\newblock \bibinfo{journal}{IEEE Access} \bibinfo{volume}{6}
  (\bibinfo{year}{2018}) \bibinfo{pages}{52138--52160}.
\bibitem[{Guidotti et~al.(2018)Guidotti, Monreale, Ruggieri, Turini, Giannotti,
  and Pedreschi}]{Deep_explanation_survey}
\bibinfo{author}{R.~Guidotti}, \bibinfo{author}{A.~Monreale},
  \bibinfo{author}{S.~Ruggieri}, \bibinfo{author}{F.~Turini},
  \bibinfo{author}{F.~Giannotti}, \bibinfo{author}{D.~Pedreschi},
\newblock \bibinfo{title}{A survey of methods for explaining black box models},
\newblock \bibinfo{journal}{ACM Comput. Surv.} \bibinfo{volume}{51}
  (\bibinfo{year}{2018}) \bibinfo{pages}{93:1--93:42}. \URLprefix
  \url{http://doi.acm.org/10.1145/3236009}. \DOIprefix\doi{10.1145/3236009}.
\bibitem[{Lowry and Macpherson(1988)}]{lowry1988blot}
\bibinfo{author}{S.~Lowry}, \bibinfo{author}{G.~Macpherson},
\newblock \bibinfo{title}{A blot on the profession},
\newblock \bibinfo{journal}{British medical journal (Clinical research ed.)}
  \bibinfo{volume}{296} (\bibinfo{year}{1988}) \bibinfo{pages}{657}.
\bibitem[{Miller(2019)}]{MILLER_19}
\bibinfo{author}{T.~Miller},
\newblock \bibinfo{title}{Explanation in artificial intelligence: Insights from
  the social sciences},
\newblock \bibinfo{journal}{Artificial Intelligence} \bibinfo{volume}{267}
  (\bibinfo{year}{2019}) \bibinfo{pages}{1 -- 38}. \URLprefix
  \url{http://www.sciencedirect.com/science/article/pii/S0004370218305988}.
  \DOIprefix\doi{https://doi.org/10.1016/j.artint.2018.07.007}.
\bibitem[{Arrieta et~al.(2019)Arrieta, Rodr{\'{\i}}guez, Ser, Bennetot, Tabik,
  Barbado, Garc{\'{\i}}a, Gil{-}Lopez, Molina, Benjamins, Chatila, and
  Herrera}]{arrieta_et_al_2019}
\bibinfo{author}{A.~B. Arrieta}, \bibinfo{author}{N.~D. Rodr{\'{\i}}guez},
  \bibinfo{author}{J.~D. Ser}, \bibinfo{author}{A.~Bennetot},
  \bibinfo{author}{S.~Tabik}, \bibinfo{author}{A.~Barbado},
  \bibinfo{author}{S.~Garc{\'{\i}}a}, \bibinfo{author}{S.~Gil{-}Lopez},
  \bibinfo{author}{D.~Molina}, \bibinfo{author}{R.~Benjamins},
  \bibinfo{author}{R.~Chatila}, \bibinfo{author}{F.~Herrera},
\newblock \bibinfo{title}{Explainable artificial intelligence {(XAI):}
  concepts, taxonomies, opportunities and challenges toward responsible {AI}},
\newblock \bibinfo{journal}{CoRR} \bibinfo{volume}{abs/1910.10045}
  (\bibinfo{year}{2019}). \URLprefix \url{http://arxiv.org/abs/1910.10045}.
  \href{http://arxiv.org/abs/1910.10045}{{\tt arXiv:1910.10045}}.
\bibitem[{Jacovi et~al.(2018)Jacovi, Sar~Shalom, and Goldberg}]{jacovi_2018}
\bibinfo{author}{A.~Jacovi}, \bibinfo{author}{O.~Sar~Shalom},
  \bibinfo{author}{Y.~Goldberg},
\newblock \bibinfo{title}{Understanding convolutional neural networks for text
  classification},
\newblock in: \bibinfo{booktitle}{Proceedings of the 2018 EMNLP Workshop
  BlackboxNLP: Analyzing and Interpreting Neural Networks for NLP},
  \bibinfo{publisher}{Association for Computational Linguistics},
  \bibinfo{year}{2018}, pp. \bibinfo{pages}{56--65}. \URLprefix
  \url{http://aclweb.org/anthology/W18-5408}.
\bibitem[{Ribeiro et~al.(2016)Ribeiro, Singh, and Guestrin}]{Ribeiro_2016}
\bibinfo{author}{M.~T. Ribeiro}, \bibinfo{author}{S.~Singh},
  \bibinfo{author}{C.~Guestrin},
\newblock \bibinfo{title}{"why should i trust you?": Explaining the predictions
  of any classifier},
\newblock in: \bibinfo{booktitle}{Proceedings of the 22Nd ACM SIGKDD
  International Conference on Knowledge Discovery and Data Mining}, KDD '16,
  \bibinfo{publisher}{ACM}, \bibinfo{address}{New York, NY, USA},
  \bibinfo{year}{2016}, pp. \bibinfo{pages}{1135--1144}. \URLprefix
  \url{http://doi.acm.org/10.1145/2939672.2939778}.
  \DOIprefix\doi{10.1145/2939672.2939778}.
\bibitem[{Zhang and Wallace(2015)}]{zhang2015sensitivity}
\bibinfo{author}{Y.~Zhang}, \bibinfo{author}{B.~Wallace},
\newblock \bibinfo{title}{A sensitivity analysis of (and practitioners' guide
  to) convolutional neural networks for sentence classification},
\newblock \bibinfo{journal}{arXiv preprint arXiv:1510.03820}
  (\bibinfo{year}{2015}).
\bibitem[{Karpathy et~al.(2015)Karpathy, Johnson, and
  Fei-Fei}]{karpathy2015visualizing}
\bibinfo{author}{A.~Karpathy}, \bibinfo{author}{J.~Johnson},
  \bibinfo{author}{L.~Fei-Fei},
\newblock \bibinfo{title}{Visualizing and understanding recurrent networks},
\newblock \bibinfo{journal}{arXiv preprint arXiv:1506.02078}
  (\bibinfo{year}{2015}).
\bibitem[{Bach et~al.(2015)Bach, Binder, Montavon, Klauschen, Müller, and
  Samek}]{Bach_LRP}
\bibinfo{author}{S.~Bach}, \bibinfo{author}{A.~Binder},
  \bibinfo{author}{G.~Montavon}, \bibinfo{author}{F.~Klauschen},
  \bibinfo{author}{K.-R. Müller}, \bibinfo{author}{W.~Samek},
\newblock \bibinfo{title}{On pixel-wise explanations for non-linear classifier
  decisions by layer-wise relevance propagation},
\newblock \bibinfo{journal}{PLOS ONE} \bibinfo{volume}{10}
  (\bibinfo{year}{2015}) \bibinfo{pages}{1--46}. \URLprefix
  \url{https://doi.org/10.1371/journal.pone.0130140}.
  \DOIprefix\doi{10.1371/journal.pone.0130140}.
\bibitem[{Carter et~al.(2019)Carter, Mueller, Jain, and
  Gifford}]{carter2019made}
\bibinfo{author}{B.~Carter}, \bibinfo{author}{J.~Mueller},
  \bibinfo{author}{S.~Jain}, \bibinfo{author}{D.~Gifford},
\newblock \bibinfo{title}{What made you do this? understanding black-box
  decisions with sufficient input subsets},
\newblock in: \bibinfo{booktitle}{The 22nd International Conference on
  Artificial Intelligence and Statistics}, \bibinfo{organization}{PMLR},
  \bibinfo{year}{2019}, pp. \bibinfo{pages}{567--576}.
\bibitem[{Tjoa and Guan(2019)}]{Tjoa2019_survey}
\bibinfo{author}{E.~Tjoa}, \bibinfo{author}{C.~Guan},
\newblock \bibinfo{title}{A survey on explainable artificial intelligence
  (xai): Towards medical xai},
\newblock \bibinfo{journal}{ArXiv} \bibinfo{volume}{abs/1907.07374}
  (\bibinfo{year}{2019}).
\bibitem[{Zeiler~Matthew and Rob(2013)}]{zeiler2013visualizing}
\bibinfo{author}{D.~Zeiler~Matthew}, \bibinfo{author}{F.~Rob},
\newblock \bibinfo{title}{Visualizing and understanding convolutional
  networks},
\newblock \bibinfo{journal}{CoRR.----2013.----Vol. abs/1311.2901.----URL:
  http://arxiv. org/abs/1311.2901}  (\bibinfo{year}{2013}).
\bibitem[{Kim et~al.(2018)Kim, Wattenberg, Gilmer, Cai, Wexler, Viegas
  et~al.}]{kim2018interpretability}
\bibinfo{author}{B.~Kim}, \bibinfo{author}{M.~Wattenberg},
  \bibinfo{author}{J.~Gilmer}, \bibinfo{author}{C.~Cai},
  \bibinfo{author}{J.~Wexler}, \bibinfo{author}{F.~Viegas}, et~al.,
\newblock \bibinfo{title}{Interpretability beyond feature attribution:
  Quantitative testing with concept activation vectors (tcav)},
\newblock in: \bibinfo{booktitle}{International conference on machine
  learning}, \bibinfo{organization}{PMLR}, \bibinfo{year}{2018}, pp.
  \bibinfo{pages}{2668--2677}.
\bibitem[{Tishby and Zaslavsky(2015)}]{tishby2015deep}
\bibinfo{author}{N.~Tishby}, \bibinfo{author}{N.~Zaslavsky},
\newblock \bibinfo{title}{Deep learning and the information bottleneck
  principle},
\newblock in: \bibinfo{booktitle}{2015 IEEE Information Theory Workshop (ITW)},
  \bibinfo{organization}{IEEE}, \bibinfo{year}{2015}, pp.
  \bibinfo{pages}{1--5}.
\bibitem[{Shwartz-Ziv and Tishby(2017)}]{shwartz2017opening}
\bibinfo{author}{R.~Shwartz-Ziv}, \bibinfo{author}{N.~Tishby},
\newblock \bibinfo{title}{Opening the black box of deep neural networks via
  information},
\newblock \bibinfo{journal}{arXiv preprint arXiv:1703.00810}
  (\bibinfo{year}{2017}).
\bibitem[{{Selvaraju} et~al.(2017){Selvaraju}, {Cogswell}, {Das}, {Vedantam},
  {Parikh}, and {Batra}}]{Selvaraju_17}
\bibinfo{author}{R.~R. {Selvaraju}}, \bibinfo{author}{M.~{Cogswell}},
  \bibinfo{author}{A.~{Das}}, \bibinfo{author}{R.~{Vedantam}},
  \bibinfo{author}{D.~{Parikh}}, \bibinfo{author}{D.~{Batra}},
\newblock \bibinfo{title}{Grad-cam: Visual explanations from deep networks via
  gradient-based localization},
\newblock in: \bibinfo{booktitle}{2017 IEEE International Conference on
  Computer Vision (ICCV)}, \bibinfo{year}{2017}, pp. \bibinfo{pages}{618--626}.
  \DOIprefix\doi{10.1109/ICCV.2017.74}.
\bibitem[{Liu et~al.(2020)Liu, Melkman, and Akutsu}]{LiuMA20}
\bibinfo{author}{P.~Liu}, \bibinfo{author}{A.~A. Melkman},
  \bibinfo{author}{T.~Akutsu},
\newblock \bibinfo{title}{Extracting boolean and probabilistic rules from
  trained neural networks},
\newblock \bibinfo{journal}{Neural Networks} \bibinfo{volume}{126}
  (\bibinfo{year}{2020}) \bibinfo{pages}{300--311}. \URLprefix
  \url{https://doi.org/10.1016/j.neunet.2020.03.024}.
  \DOIprefix\doi{10.1016/j.neunet.2020.03.024}.
\bibitem[{Ghorbani et~al.(2019)Ghorbani, Wexler, Zou, and
  Kim}]{ghorbani2019towards}
\bibinfo{author}{A.~Ghorbani}, \bibinfo{author}{J.~Wexler},
  \bibinfo{author}{J.~Y. Zou}, \bibinfo{author}{B.~Kim},
\newblock \bibinfo{title}{Towards automatic concept-based explanations},
\newblock in: \bibinfo{booktitle}{Advances in Neural Information Processing
  Systems}, \bibinfo{year}{2019}, pp. \bibinfo{pages}{9277--9286}.
\bibitem[{Shrikumar et~al.(2016)Shrikumar, Greenside, Shcherbina, and
  Kundaje}]{shrikumar2016not}
\bibinfo{author}{A.~Shrikumar}, \bibinfo{author}{P.~Greenside},
  \bibinfo{author}{A.~Shcherbina}, \bibinfo{author}{A.~Kundaje},
\newblock \bibinfo{title}{Not just a black box: Learning important features
  through propagating activation differences},
\newblock \bibinfo{journal}{arXiv preprint arXiv:1605.01713}
  (\bibinfo{year}{2016}).
\bibitem[{Zintgraf et~al.(2017)Zintgraf, Cohen, Adel, and
  Welling}]{zintgraf2017visualizing}
\bibinfo{author}{L.~M. Zintgraf}, \bibinfo{author}{T.~S. Cohen},
  \bibinfo{author}{T.~Adel}, \bibinfo{author}{M.~Welling},
\newblock \bibinfo{title}{Visualizing deep neural network decisions: Prediction
  difference analysis},
\newblock \bibinfo{journal}{arXiv preprint arXiv:1702.04595}
  (\bibinfo{year}{2017}).
\bibitem[{Kindermans et~al.(2017)Kindermans, Sch{\"u}tt, Alber, M{\"u}ller,
  Erhan, Kim, and D{\"a}hne}]{kindermans2017learning}
\bibinfo{author}{P.-J. Kindermans}, \bibinfo{author}{K.~T. Sch{\"u}tt},
  \bibinfo{author}{M.~Alber}, \bibinfo{author}{K.-R. M{\"u}ller},
  \bibinfo{author}{D.~Erhan}, \bibinfo{author}{B.~Kim},
  \bibinfo{author}{S.~D{\"a}hne},
\newblock \bibinfo{title}{Learning how to explain neural networks: Patternnet
  and patternattribution},
\newblock \bibinfo{journal}{arXiv preprint arXiv:1705.05598}
  (\bibinfo{year}{2017}).
\bibitem[{Angelov and Soares(2020)}]{plamen_20}
\bibinfo{author}{P.~Angelov}, \bibinfo{author}{E.~Soares},
\newblock \bibinfo{title}{Towards explainable deep neural networks (xdnn)},
\newblock \bibinfo{journal}{Neural Networks} \bibinfo{volume}{130}
  (\bibinfo{year}{2020}). \DOIprefix\doi{10.1016/j.neunet.2020.07.010}.
\bibitem[{Erhan et~al.(2009)Erhan, Bengio, Courville, and
  Vincent}]{erhan2009visualizing}
\bibinfo{author}{D.~Erhan}, \bibinfo{author}{Y.~Bengio},
  \bibinfo{author}{A.~Courville}, \bibinfo{author}{P.~Vincent},
  \bibinfo{title}{Visualizing Higher-Layer Features of a Deep Network},
  \bibinfo{type}{Technical Report}, Univeristy of Montreal,
  \bibinfo{year}{2009}.
\bibitem[{Szegedy et~al.(2015)Szegedy, Liu, Jia, Sermanet, Reed, Anguelov,
  Erhan, Vanhoucke, and Rabinovich}]{szegedy2015going}
\bibinfo{author}{C.~Szegedy}, \bibinfo{author}{W.~Liu},
  \bibinfo{author}{Y.~Jia}, \bibinfo{author}{P.~Sermanet},
  \bibinfo{author}{S.~Reed}, \bibinfo{author}{D.~Anguelov},
  \bibinfo{author}{D.~Erhan}, \bibinfo{author}{V.~Vanhoucke},
  \bibinfo{author}{A.~Rabinovich},
\newblock \bibinfo{title}{Going deeper with convolutions},
\newblock in: \bibinfo{booktitle}{Proceedings of the IEEE conference on
  computer vision and pattern recognition}, \bibinfo{year}{2015}, pp.
  \bibinfo{pages}{1--9}.
\bibitem[{Kindermans et~al.(2019)Kindermans, Hooker, Adebayo, Alber,
  Sch{\"u}tt, D{\"a}hne, Erhan, and Kim}]{kindermans2019reliability}
\bibinfo{author}{P.-J. Kindermans}, \bibinfo{author}{S.~Hooker},
  \bibinfo{author}{J.~Adebayo}, \bibinfo{author}{M.~Alber},
  \bibinfo{author}{K.~T. Sch{\"u}tt}, \bibinfo{author}{S.~D{\"a}hne},
  \bibinfo{author}{D.~Erhan}, \bibinfo{author}{B.~Kim},
\newblock \bibinfo{title}{The (un) reliability of saliency methods},
\newblock in: \bibinfo{booktitle}{Explainable AI: Interpreting, Explaining and
  Visualizing Deep Learning}, \bibinfo{publisher}{Springer},
  \bibinfo{year}{2019}, pp. \bibinfo{pages}{267--280}.
\bibitem[{Caruana et~al.(2015)Caruana, Lou, Gehrke, Koch, Sturm, and
  Elhadad}]{caruana2015intelligible}
\bibinfo{author}{R.~Caruana}, \bibinfo{author}{Y.~Lou},
  \bibinfo{author}{J.~Gehrke}, \bibinfo{author}{P.~Koch},
  \bibinfo{author}{M.~Sturm}, \bibinfo{author}{N.~Elhadad},
\newblock \bibinfo{title}{Intelligible models for healthcare: Predicting
  pneumonia risk and hospital 30-day readmission},
\newblock in: \bibinfo{booktitle}{Proceedings of the 21th ACM SIGKDD
  international conference on knowledge discovery and data mining},
  \bibinfo{year}{2015}, pp. \bibinfo{pages}{1721--1730}.
\bibitem[{Letham et~al.(2015)Letham, Rudin, McCormick, Madigan
  et~al.}]{letham2015interpretable}
\bibinfo{author}{B.~Letham}, \bibinfo{author}{C.~Rudin}, \bibinfo{author}{T.~H.
  McCormick}, \bibinfo{author}{D.~Madigan}, et~al.,
\newblock \bibinfo{title}{Interpretable classifiers using rules and bayesian
  analysis: Building a better stroke prediction model},
\newblock \bibinfo{journal}{The Annals of Applied Statistics}
  \bibinfo{volume}{9} (\bibinfo{year}{2015}) \bibinfo{pages}{1350--1371}.
\bibitem[{Lapuschkin et~al.(2019)Lapuschkin, W{\"a}ldchen, Binder, Montavon,
  Samek, and M{\"u}ller}]{lapuschkin2019unmasking}
\bibinfo{author}{S.~Lapuschkin}, \bibinfo{author}{S.~W{\"a}ldchen},
  \bibinfo{author}{A.~Binder}, \bibinfo{author}{G.~Montavon},
  \bibinfo{author}{W.~Samek}, \bibinfo{author}{K.-R. M{\"u}ller},
\newblock \bibinfo{title}{Unmasking clever hans predictors and assessing what
  machines really learn},
\newblock \bibinfo{journal}{Nature communications} \bibinfo{volume}{10}
  (\bibinfo{year}{2019}) \bibinfo{pages}{1--8}.
\bibitem[{Samek et~al.(2016)Samek, Binder, Montavon, Lapuschkin, and
  M{\"u}ller}]{samek2016evaluating}
\bibinfo{author}{W.~Samek}, \bibinfo{author}{A.~Binder},
  \bibinfo{author}{G.~Montavon}, \bibinfo{author}{S.~Lapuschkin},
  \bibinfo{author}{K.-R. M{\"u}ller},
\newblock \bibinfo{title}{Evaluating the visualization of what a deep neural
  network has learned},
\newblock \bibinfo{journal}{IEEE transactions on neural networks and learning
  systems} \bibinfo{volume}{28} (\bibinfo{year}{2016})
  \bibinfo{pages}{2660--2673}.
\bibitem[{Arras et~al.(2017)Arras, Horn, Montavon, M{\"u}ller, and
  Samek}]{arras2017relevant}
\bibinfo{author}{L.~Arras}, \bibinfo{author}{F.~Horn},
  \bibinfo{author}{G.~Montavon}, \bibinfo{author}{K.-R. M{\"u}ller},
  \bibinfo{author}{W.~Samek},
\newblock \bibinfo{title}{" what is relevant in a text document?": An
  interpretable machine learning approach},
\newblock \bibinfo{journal}{PloS one} \bibinfo{volume}{12}
  (\bibinfo{year}{2017}) \bibinfo{pages}{e0181142}.
\bibitem[{Kim(2014)}]{kim-2014-convolutional}
\bibinfo{author}{Y.~Kim},
\newblock \bibinfo{title}{Convolutional neural networks for sentence
  classification},
\newblock in: \bibinfo{booktitle}{Proceedings of the 2014 Conference on
  Empirical Methods in Natural Language Processing ({EMNLP})},
  \bibinfo{publisher}{Association for Computational Linguistics},
  \bibinfo{address}{Doha, Qatar}, \bibinfo{year}{2014}, pp.
  \bibinfo{pages}{1746--1751}. \URLprefix
  \url{https://www.aclweb.org/anthology/D14-1181}.
  \DOIprefix\doi{10.3115/v1/D14-1181}.
\bibitem[{Jiechieu and Tsopze(2020)}]{jiechieu2020skills}
\bibinfo{author}{K.~F.~F. Jiechieu}, \bibinfo{author}{N.~Tsopze},
\newblock \bibinfo{title}{Skills prediction based on multi-label resume
  classification using cnn with model predictions explanation},
\newblock \bibinfo{journal}{Neural Computing and Applications}
  (\bibinfo{year}{2020}) \bibinfo{pages}{1--19}.
\bibitem[{Maas et~al.(2011)Maas, Daly, Pham, Huang, Ng, and
  Potts}]{maas-EtAl:2011:ACL-HLT2011}
\bibinfo{author}{A.~L. Maas}, \bibinfo{author}{R.~E. Daly},
  \bibinfo{author}{P.~T. Pham}, \bibinfo{author}{D.~Huang},
  \bibinfo{author}{A.~Y. Ng}, \bibinfo{author}{C.~Potts},
\newblock \bibinfo{title}{Learning word vectors for sentiment analysis},
\newblock in: \bibinfo{booktitle}{Proceedings of the 49th Annual Meeting of the
  Association for Computational Linguistics: Human Language Technologies},
  \bibinfo{publisher}{Association for Computational Linguistics},
  \bibinfo{address}{Portland, Oregon, USA}, \bibinfo{year}{2011}, pp.
  \bibinfo{pages}{142--150}. \URLprefix
  \url{http://www.aclweb.org/anthology/P11-1015}.
\bibitem[{Go et~al.(2009)Go, Bhayani, and Huang}]{go2009twitter}
\bibinfo{author}{A.~Go}, \bibinfo{author}{R.~Bhayani},
  \bibinfo{author}{L.~Huang},
\newblock \bibinfo{title}{Twitter sentiment classification using distant
  supervision},
\newblock \bibinfo{journal}{CS224N project report, Stanford}
  \bibinfo{volume}{1} (\bibinfo{year}{2009}) \bibinfo{pages}{2009}.
\bibitem[{Voorhees(2001)}]{qa_trac}
\bibinfo{author}{E.~Voorhees},
\newblock \bibinfo{title}{The trec question answering track},
\newblock \bibinfo{journal}{Nat. Lang. Eng.} \bibinfo{volume}{7}
  (\bibinfo{year}{2001}) \bibinfo{pages}{361--378}.
  \DOIprefix\doi{10.1017/S1351324901002789}.
\bibitem[{Kotzias et~al.(2015)Kotzias, Denil, De~Freitas, and
  Smyth}]{kotzias2015group}
\bibinfo{author}{D.~Kotzias}, \bibinfo{author}{M.~Denil},
  \bibinfo{author}{N.~De~Freitas}, \bibinfo{author}{P.~Smyth},
\newblock \bibinfo{title}{From group to individual labels using deep features},
\newblock in: \bibinfo{booktitle}{Proceedings of the 21th ACM SIGKDD
  International Conference on Knowledge Discovery and Data Mining},
  \bibinfo{year}{2015}, pp. \bibinfo{pages}{597--606}.

\end{thebibliography}







\end{document}